\documentclass{egpubl}

\ConferenceSubmission 
\electronicVersion

\ifpdf \usepackage[pdftex]{graphicx} \pdfcompresslevel=9
\else \usepackage[dvips]{graphicx} \fi

\PrintedOrElectronic

\usepackage{bm,overpic}
\usepackage{t1enc,dfadobe}
\usepackage{egweblnk}
\usepackage{cite}

\usepackage{etex}

\usepackage[english]{babel}
\usepackage[utf8]{inputenc}
\usepackage{moresize}
\usepackage{amsmath,amssymb}
\usepackage{graphics,graphicx}
\usepackage{hyperref}
\usepackage{tabularx}
\usepackage{booktabs} 
\usepackage{caption} 
\usepackage{multirow}

\usepackage{color}
\definecolor{myred}{rgb}{0.75,0,0}
\definecolor{mygreen}{rgb}{0,0.5,0}
\definecolor{myblue}{rgb}{0,0,0.75}
\definecolor{dblp_blue}{rgb}{0.25,0.42,0.63}
\definecolor{dblp_red}{rgb}{0.67,0.18,0.44}
\definecolor{dblp_gray}{rgb}{0.39,0.42,0.44}

\usepackage{pgf,tikz}
\usepackage{tkz-euclide}
\usepackage{tkz-graph}
\usetikzlibrary{calc}
\usetkzobj{all}

\usepackage{pgfplots}
\newlength\figureheight 
\newlength\figurewidth 

\usepackage{overpic}

\usepackage{fixmath}

\newcommand{\etal}{\emph{et al.~}}
\newcommand{\T}{\ensuremath{^\top}}
\newcommand{\vct}[1]{\ensuremath{\mathbf{#1}}}

\newenvironment{psmallmatrix}
  {\left(\begin{smallmatrix}}
  {\end{smallmatrix}\right)}

\title[Learning correspondence with anisotropic CNN]{Learning shape correspondence with \\ anisotropic convolutional neural networks}

\author[D. Boscaini et al.]
	{Davide Boscaini$^1$, Jonathan Masci$^1$, Emanuele Rodol\`a$^1$, and Michael M. Bronstein$^{1,2}$\\
	$^1$Institute of Computational Science, Faculty of Informatics, University of Lugano (USI), Switzerland\\
	$^2$Perceptual Computing, Intel, Israel}
       
\begin{document}

 \teaser{
 	\vspace*{-1.25mm}
	\includegraphics[width=\linewidth]{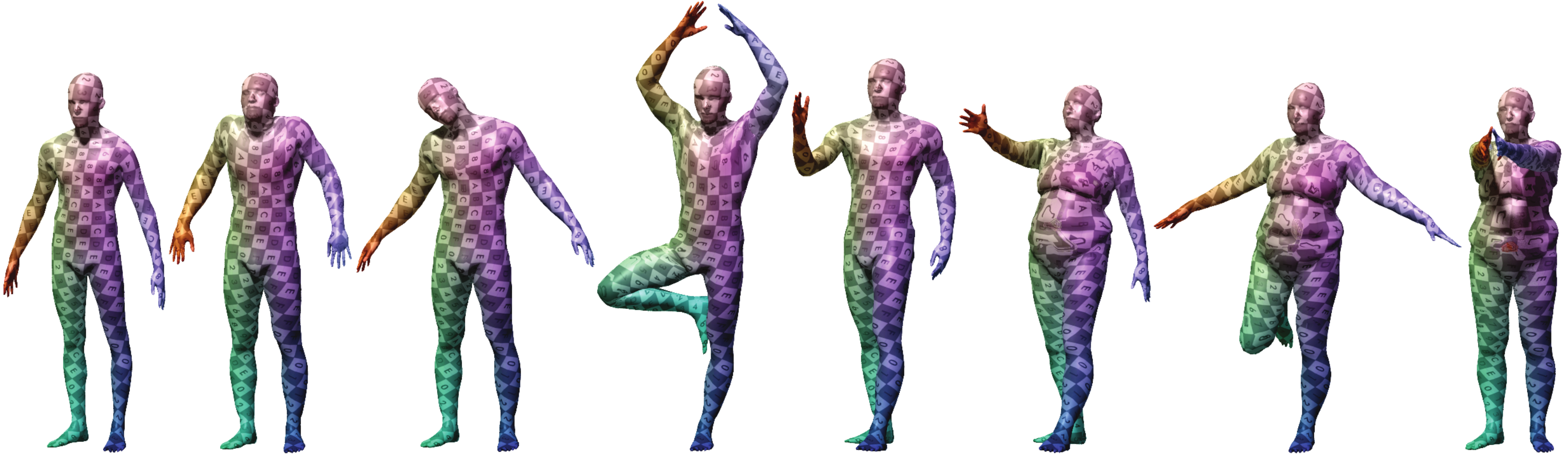}
 	\centering
 	\caption{
	Examples of correspondence on the FAUST humans dataset obtained by the proposed ACNN method. Shown is the texture transferred from the leftmost reference shape to different subjects in different poses by means of our correspondence. The correspondence is nearly perfect (only very few minor artifacts are noticeable). See text for details. 
	\smallskip
	}
	\label{fig:teaser}
}

\maketitle

\begin{abstract}

Establishing correspondence between shapes is a fundamental problem in geometry processing, arising in a wide variety of applications. The problem is especially difficult in the setting of non-isometric deformations, as well as in the presence of topological noise and missing parts, mainly due to the limited capability to model such deformations axiomatically. 
Several recent works showed that invariance to complex shape transformations can be learned from examples. 
In this paper, we introduce an intrinsic convolutional neural network architecture based on anisotropic diffusion kernels, which we term Anisotropic Convolutional Neural Network (ACNN). 
%
%
In our construction, we generalize convolutions to non-Euclidean domains by constructing a set of oriented anisotropic diffusion kernels, creating in this way a local intrinsic polar representation of the data (`patch'), which is then correlated with a filter.  
Several cascades of such filters, linear, and non-linear operators are stacked to form a deep neural network whose parameters are learned by minimizing a task-specific cost.
We use ACNNs to effectively learn intrinsic dense correspondences between deformable shapes in very challenging settings, achieving state-of-the-art results on some of the most difficult recent correspondence benchmarks.

\end{abstract}

\section{Introduction}

In geometry processing, computer graphics, and vision, finding intrinsic correspondence between 3D shapes affected by different transformations is one of the fundamental problems with a wide spectrum of applications ranging from texture mapping to animation \cite{kaick2010survey}. 
Of particular interest is the setting in which the shapes are allowed to deform non-rigidly. Recently, the emergence of 3D sensing technology has brought the need to deal with acquisition artifacts, such as missing parts, geometric, and topological noise, as well as matching 3D shapes in different representations, such as meshes and point clouds. 
The main topic of this paper is establishing dense intrinsic correspondence between non-rigid shapes in such challenging settings.

\subsection{Related works} 

\paragraph*{Correspondence.}
Traditional correspondence approaches try to find a {\em point-wise} matching between (a subset of) the points on two or more shapes. 
\emph{Minimum-distortion methods} establish the matching by minimizing some structure distortion, which can include similarity of local features \cite{OvMe*10,WKS,ZaBo*09}, geodesic \cite{Memoli:2005,bro:bro:kim:PNAS,koltun} or diffusion distances \cite{lafon:05:LOCAL}, or a combination thereof \cite{torresani2008feature}, or higher-order structures \cite{zeng2010dense}. 
Typically, the computational complexity of such methods is high, and there have been several attempts to alleviate the computational complexity using hierarchical \cite{sahillioglu2012} or subsampling \cite{tevs2011intrinsic} methods. 
Several approaches formulate the correspondence problem as quadratic assignment and employ different relaxations thereof \cite{umeyama1988eigendecomposition,leordeanu2005spectral,rodola2012game,koltun,lipman15}.

\emph{Embedding methods} try to exploit some assumption on the shapes (e.g. approximate isometry) in order to parametrize the correspondence problem with a small number of degrees of freedom. 
Elad and Kimmel \cite{ela:kim:FLATTEN} used multi-dimensional scaling to embed the geodesic metric of the matched shapes into a low-dimensional Euclidean space, where alignment of the resulting ``canonical forms'' is then performed by simple rigid matching (ICP) \cite{ChenMedioni:91:ICP,bes:mck:SURFACEMATCH}. 
The works of \cite{Mateus08,shtern2013matching} used the eigenfunctions of the Laplace-Beltrami operator as embedding coordinates and performed matching in the eigenspace. 
Lipman \etal \cite{Lipman2011,KimLCF10,kim2011blended} used conformal embeddings into disks and spheres to parametrize correspondences between homeomorphic surfaces as M{\"o}bius transformations.

As opposed to point-wise correspondence methods, \emph{soft correspondence} approaches assign a point on one shape to more than one point on the other. 
Several methods formulated soft correspondence as a mass-transportation problem \cite{Me11,solomon2012soft}. 
Ovsjanikov \etal \cite{ovsjanikov2012functional} introduced the \emph{functional correspondence} framework, modeling the correspondence as a linear operator between spaces of functions on two shapes, which has an efficient representation in the Laplacian eigenbases. 
This approach was extended in several follow-up works \cite{pokrass2013sparse,kovnatsky15,SGMDS,PFM2016} .

In the past year, we have witnessed the emergence of \emph{learning-based approaches} for 3D shape correspondence. 
The dramatic success of deep learning (in particular, convolutional neural networks \cite{fukushima1980neocognitron,lecun1989backpropagation}) in computer vision \cite{Krizhevsky:2012}  has lead to a recent keen interest in the geometry processing and graphics communities to apply such methodologies to geometric problems \cite{masci2015shapenet,kalogerakis2015,Wu,WFT2015,Li}.

\begin{figure}[t]
\centering
\begin{overpic}
	[width=1\linewidth]{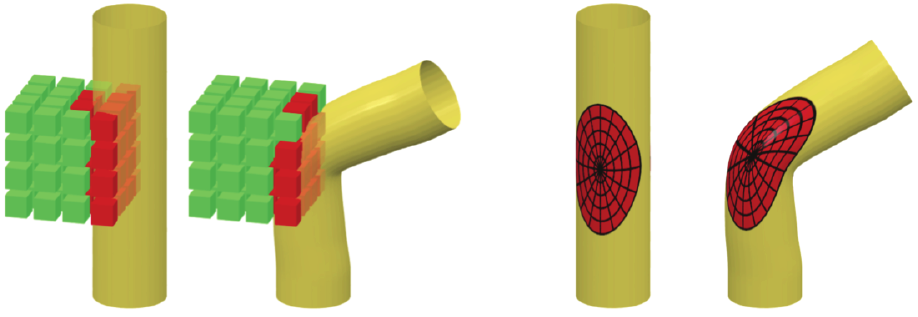}
\end{overpic}
\caption{
Illustration of the difference between extrinsic and intrinsic deep learning methods on geometric data. 
Left: extrinsic methods such as volumetric CNNs treat 3D geometric data in its Euclidean representation. Such a representation is not invariant to deformations (e.g., in the shown example, the filter that responds to features on a straight cylinder would not respond to a bent one). Right: in an intrinsic representation, the filter is applied to some data on the surface itself, thus being invariant to deformations. 
}
\label{fig:extvsint}
\end{figure}

\paragraph*{Extrinsic deep learning.} 
Many machine learning techniques successfully working on images were tried ``as is'' on 3D geometric data, represented for this purpose in some way ``digestible'' by standard frameworks. 
Su \etal \cite{kalogerakis2015} used CNNs applied to range images obtained from multiple views of 3D objects for retrieval and classification tasks. 
Wei \etal \cite{Li} used view-based representation to find correspondence between non-rigid shapes. 
Wu \etal \cite{Wu} used volumetric CNNs applied to rasterized volumetric representation of 3D shapes. 

The main drawback of such approaches is their treatment of geometric data as Euclidean structures (see Figure~\ref{fig:extvsint}). First, for complex 3D objects, Euclidean representations such as depth images or voxels may lose significant parts of the object or its fine details, or even break its topological structure (in particular, due to computational reasons, the volumetric CNNs \cite{Wu} used a $64\times 64\times 64$ cube, allowing only a very coarse representation of 3D geometry). Second, Euclidean representations are not intrinsic, and vary as the result of pose or deformation of the object. Achieving invariance to shape deformations, a common requirement in many applications, is extremely hard with the aforementioned methods and requires complex models and huge training sets due to the large number of degrees of freedom involved in describing non-rigid deformations.

\paragraph*{Intrinsic deep learning} 
approaches try to apply learning techniques to geometric data by generalizing the main ingredients such as convolutions to non-Euclidean domains. 
One of the first attempts to learn spectral kernels for shape recognition was done in \cite{Aflalo}. Litman and Bronstein \cite{OSD} learned optimal spectral descriptors that generalize the popular ``handcrafted'' heat- \cite{HKS1,HKS2} and wave-kernel signatures \cite{WKS} and show performance superior to both. 
Their construction was recently extended in \cite{ADD} using anisotropic spectral kernels, referred to as Anisotropic Diffusion Descriptors (ADD), based on the anisotropic Laplace-Beltrami operator \cite{albo}. A key advantage of the resulting approach is the ability to disambiguate intrinsic symmetries \cite{ovsjanikov2008global}, to which most of the standard spectral descriptors are agnostic. 
Corman \etal \cite{corman2014supervised} used descriptor learning in the functional maps framework.
Rodol{\`a} \etal \cite{rodoladense} proposed learning correspondences between non-rigid shapes using random forests applied to WKS descriptors.

The first intrinsic convolutional neural network architecture (Geodesic CNN) was presented in \cite{masci2015shapenet}. GCNN is based on a local intrinsic charting procedure from \cite{ISC}, 
and while producing impressive results on several shape correspondence and retrieval benchmarks, has a number of significant drawbacks. First, the charting procedure is limited to meshes, and second, there is no guarantee that the chart is always topologically meaningful. 

Another intrinsic CNN construction (Localized Spectral CNN) using an alternative charting technique based on the windowed Fourier transform \cite{Shumann} was proposed in \cite{WFT2015}. This method is a generalization of a previous work \cite{Bruna} on spectral deep learning on graphs. One of the key advantages of LSCNN is that the same framework can be applied to different shape representations, in particular, meshes and point clouds. 
A drawback of this approach is its memory and computation requirements, as each window needs to be explicitly produced.

\subsection{Main contributions} 
In this paper, we present Anisotropic Convolutional Neural Networks (ACNN), a method for intrinsic deep learning on non-Euclidean domains. Though it is a generic framework that can be used to handle different tasks, we focus here on learning correspondence between shapes.

Our approach is related to two previous methods for deep learning on manifolds, GCNN \cite{masci2015shapenet} and ADD \cite{ADD}. 
Compared to \cite{ADD}, where a learned spectral filter applied to the eigenvalues of anisotropic Laplace-Beltrami operator, we use anisotropic heat kernels as spatial weighting functions allowing to extract a local intrinsic representation of a function defined on the manifold. Unlike ADD, our ACNN is a convolutional neural network architecture. 
Compared to GCNN, our construction of the ``patch operator'' is much simpler, does not depend on the injectivity radius of the manifold, and is not limited to triangular meshes. 
Overall, ACNN combines all the best properties of the previous approaches without inheriting their drawbacks. 
We show that the proposed  framework beats GCNN, ADD, and other state-of-the-art approaches on challenging correspondence benchmarks.

The rest of the paper is organized as follows. 
In Section~\ref{sec:background} we overview the main notions related to spectral analysis on manifolds and define anisotropic Laplacians and heat kernels. 
In Section~\ref{sec:deeplearn} we briefly discuss previous approaches to intrinsic deep learning on manifolds and their drawbacks. 
Section~\ref{sec:our} describes the proposed ACNN construction for learning intrinsic dense correspondence between shapes. 
Section~\ref{sec:discrete} discussed the discretization and numerical implementation. 
In Section~\ref{sec:results} we evaluate the proposed approach on standard benchmarks and compare it to previous state-of-the-art methods. 
Finally, Section~\ref{sec:concl} concludes the paper.

\section{Background}
\label{sec:background}

\subsection{Basic notions}

\paragraph*{Manifolds.}
We model a $3$D shape as a two-dimensional compact Riemannian manifold (surface) $X$, possibly with boundary $\partial X$. 
Let $T_x X$ denote the \emph{tangent plane} at $x$, modeling the surface locally as a Euclidean space, $TX$ denote the \emph{tangent bundle}, and let 
 $\text{exp}_x \colon T_x X \to X$ be the \emph{exponential map}, mapping tangent vectors onto the surface. 
A \emph{Riemannian metric} is an inner product $\langle \cdot,\cdot \rangle_{T_x X} \colon T_x X \times T_x X \to \mathbb{R}$ on the tangent plane, depending smoothly on $x$. The Riemannian metric is represented as a $2\times 2$ matrix referred to as {\em first fundamental form}. 
Quantities which are expressible entirely in terms of Riemannian metric, and therefore independent on the way the surface is embedded, are called \emph{intrinsic}. Such quantities are invariant to isometric (metric-preserving) deformations.

\paragraph*{Curvature.}
Given an embedding of the surface, the \emph{second fundamental form}, represented as a $2\times 2$ matrix at each point, describes how the surface locally differs from a plane. 
The eigenvalues $\kappa_m, \kappa_M$ of the second fundamental form are called the \emph{principal curvatures}; the corresponding eigenvectors $\vct{v}_m, \vct{v}_M$ called the \emph{principal curvature directions} form an orthonormal basis on the tangent plane.

\paragraph*{Differential operators on manifolds.} 
Let $f \colon X \to \mathbb{R}$ be a smooth scalar field on the surface. The \emph{intrinsic gradient} is defined as 
\[
\nabla_X f(x) = \nabla (f \circ \exp_x)(\mathbf{0}), 
\]
where $\nabla$ denotes the standard Euclidean gradient acting in the tangent plane. 
The intrinsic gradient can be interpreted as the direction (tangent vector on $T_x X$) in which $f$ changes the most at point $x$. 
First-order Taylor expansion takes the form  $(f \circ \exp_x)(\mathbf{v}) \approx f(x) + \langle \nabla_X f(x), \mathbf{v} \rangle_{T_x X}$, where the second term is the {\em directional derivative} of $f$ along the tangent vector $\mathbf{v} \in T_x X$. 

Given a smooth vector field $\mathbf{v} \colon X \to TX$, the \emph{intrinsic divergence} is an operator acting on vector fields producing scalar fields, defined as the negative adjoint of the intrinsic gradient operator, 
\begin{equation}
\int_X \langle \nabla_X f(x), \mathbf{v}(x) \rangle_{T_x X} dx = - \int_X f(x)  \text{div}_X \mathbf{v}(x) dx,  
\end{equation}
where the area element $dx$ is induced by the Riemannian metric. 
Combining the two, we define the \emph{Laplace-Beltrami operator} as
\begin{equation}\label{eq:LBO}
\Delta_X f(x) = - \text{div}_X ( \nabla_X f(x) ). 
\end{equation}

\subsection{Spectral analysis on manifolds}

\paragraph*{Laplacian eigenfunctions and eigenvalues.}
$\Delta_X$ is a positive-semidefinite operator, admitting real eigendecomposition
\begin{eqnarray}
\Delta_{X} \phi_i(x) = \lambda_i \phi_i(x)  & \quad & x \in \mathrm{int}(X) \\
\langle \nabla_{X} \phi_i(x) , \hat{n}(x) \rangle = 0 & \quad & x \in \partial X, 
\label{eq:neumann}
\end{eqnarray}
with homogeneous Neumann boundary conditions~(\ref{eq:neumann}) if $X$ has a boundary  (here $\hat{n}$ denotes the normal vector to the boundary), where $0 = \lambda_0 <\lambda_1 \leq \hdots$ are eigenvalues and  $\phi_0, \phi_1, \hdots$ are the corresponding eigenfunctions.  
The eigenfunctions are orthonormal w.r.t. the standard inner product $\langle \phi_i, \phi_j \rangle_X = \int_X \phi_i(x) \phi_j(x) dx = \delta_{ij}$ and form an orthonormal basis for the functional space $L^2(X) = \{ f: X \rightarrow \mathbb{R} : \langle f, f \rangle_X  < \infty \}$.

The Laplacian eigenfunctions are a generalization of the classical Fourier basis to non-Euclidean domains: 
a function $f\in L^2(X)$ can be represented as the {\em Fourier series} 
\begin{eqnarray}
f(x) &=& \sum_{k\geq 0} \underbrace{\langle f, \phi_k \rangle_X}_{\hat{f}_k} \phi_k(x), 
\label{eq:fourier}
\end{eqnarray}
where 
the eigenvalues $\{ \lambda_k \}_{k\geq 0}$ play the role of frequencies (the first eigenvalue $\lambda_0 =0$ corresponds to a constant `DC' eigenvector) and the Fourier coefficients $\{\hat{f}_k\}_{k\geq 0}$ can be interpreted as the Fourier transform of $f$. 

\paragraph*{Heat diffusion on manifolds}
%
is governed by the {\em heat equation}, which, in the simplest case of {\em homogeneous} and {\em isotropic} heat conductivity properties of the surface, takes the form 
\begin{equation}\label{eq:heat_eq}
f_t(x,t) = -\Delta_X f(x,t),
\end{equation}
where $f(x,t)$ denotes the temperature at point $x$ at time $t$, and appropriate boundary conditions are applied if necessary. 
%
%
Given some initial heat distribution $f_0(x) = f(x,0)$, the solution of heat equation~(\ref{eq:heat_eq}) at time $t$ is obtained by applying the {\em heat operator} $H^t = e^{-t \Delta_X}$ to $f_0$, 
\begin{equation}
\label{eq:heatop}
f(x,t) = H^t f_0(x) = \int_X f_0(\xi) h_t(x,\xi)\;d\xi\,,
\end{equation}
where $h_t(x,\xi)$ is called the \emph{heat kernel}, and the above equation can be interpreted as a non-shift-invariant version of convolution.\footnote{In the Euclidean case, the heat kernel has the form $h_t(x-\xi)$ and the solution is given as $f = f_0 \ast h_t$. In signal processing terms, the heat kernel in the Euclidean case is the impulse response of a linear shift-invariant system.}  
In the spectral domain, the heat kernel is expressed as 
\begin{equation}\label{eq:heat_kernel}
h_t(x,\xi) = \sum_{k \ge 0} e^{-t \lambda_k} \phi_k(x) \phi_k(\xi). 
\end{equation}
Appealing again to the signal processing intuition, 
$e^{-t \lambda}$ acts as a low-pass filter (larger $t$ corresponding to longer diffusion results in a filter with a narrower pass band). 

\paragraph*{Spectral descriptors.}
The diagonal of the heat kernel $h_t(x,x)$, also known as {\em autodiffusivity}, for a range of values $t$, was used in \cite{HKS1,HKS2} as a local intrinsic shape descriptor referred to as the Heat Kernel Signature (HKS). The Wave Kernel Signature (WKS) \cite{WKS} uses another set of band-pass filters instead of the low-pass filters $e^{-t \lambda_k}$.  
The Optimal Spectral Descriptors (OSD) \cite{OSD} approach suggested to learn a set of optimal tasks-specific filters instead of the ``handcrafted'' low- or band-pass filters.

\subsection{Anisotropic heat kernels}

\paragraph*{Anisotropic diffusion.}
In a more general setting, the heat equation has the form 
\begin{equation}\label{eq:heat_eqa}
f_t(x,t) = -\text{div}_X(\mathbf{D}(x) \nabla_X f(x,t) ),
\end{equation}
where $\mathbf{D}(x)$ is the {\em thermal conductivity tensor} ($2\times 2$ matrix) applied to the intrinsic gradient in the tangent plane. This formulation allows modeling heat flow that is position- and direction-dependent ({\em anisotropic}). 
The special case of equation~(\ref{eq:heat_eq}) assumes $\mathbf{D}(x) = \mathbf{I}$. 

Andreux \etal \cite{albo} considered anisotropic diffusion driven by the surface curvature. Boscaini \etal \cite{ADD}, assuming that at each point $x$ the tangent vectors are expressed w.r.t. the orthogonal basis $\mathbf{v}_m, \mathbf{v}_M$ of principal curvature directions,  used a thermal conductivity tensor of the form
\begin{equation}
\label{eq:aniso_tensor}
\mathbf{D}_{\alpha \theta}(x) = \mathbf{R}_\theta(x)
\begin{bmatrix}
\alpha &  \\
 & 1
 \end{bmatrix}
\mathbf{R}^\top_\theta(x),
\end{equation}
where the $2\times 2$ matrix $\mathbf{R}_\theta(x)$ performs rotation of $\theta$ w.r.t. to the maximum curvature direction $\mathbf{v}_M(x)$, and  
$\alpha > 0$ is a parameter controlling the degree of anisotropy ($\alpha = 1$ corresponds to the classical isotropic case).

\paragraph*{Anisotropic Laplacian.}
We refer to the operator 
\[
\Delta_{\alpha\theta}f(x) = -\text{div}_X (\mathbf{D}_{\alpha \theta}(x) \nabla_X f(x)) 
\] 
as the {\em anisotropic Laplacian}, and denote by $\{ \phi_{\alpha\theta i}, \lambda_{\alpha\theta i} \}_{i\geq 0}$ its eigenfunctions and eigenvalues (computed, if applicable, with the appropriate boundary conditions).
%
Analogously to equation~(\ref{eq:heat_kernel}), the {\em anisotropic heat kernel} is given by 
\begin{equation}
\label{eq:ahks}
h_{\alpha\theta t}(x,\xi) = \sum_{k\geq 0} e^{-t \lambda_{\alpha\theta k}} \phi_{\alpha\theta k}(x) \phi_{\alpha\theta k}(\xi). 
\end{equation}
This construction was used in Anisotropic Diffusion Descriptors (ADD) \cite{ADD} to generalize the OSD approach using anisotropic heat kernels (considering the diagonal $h_{\alpha\theta t}(x,x)$ and learning a set of optimal task-specific spectral filters replacing the low-pass filters $e^{-t \lambda_{\alpha\theta k}}$).


\section{Intrinsic deep learning}
\label{sec:deeplearn}


This paper deals with the extension of convolutional neural networks (CNN) to non-Euclidean domains. 
CNN \cite{lecun1989backpropagation} have recently become extremely popular in the computer vision community due to a series of successful applications in many classically difficult problems in that domain. 
 A typical convolutional neural network architecture is hierarchical, composed of  
alternating convolutional-, pooling- (i.e. averaging), linear and non-linear layers. The parameters of different layers are learned by minimizing some task-specific cost function. The key feature of CNNs is the convolutional layer, implementing the idea of ``weight sharing'', wherein a small set of templates (filters) is applied to different parts of the data.

In image analysis applications, the input into the CNN is a function representing pixel values given on a Euclidean domain (plane); due to shift-invariance the convolution can be thought of as passing a template across the plane and recording the correlation of the template with the function at that location. 
One of the major problems in applying the CNN paradigm to non-Euclidean domains is the lack of shift-invariance, making it impossible to think of convolution as correlation with a fixed template: the template now has to be location-dependent.

There have recently been several attempts to develop intrinsic CNNs on non-Euclidean domain, which we overview below. The advantage of intrinsic CNN models over descriptor learning frameworks such as OSD \cite{OSD} or ADD \cite{ADD} is that they accept as input any information on the surface, which can represent photometric properties (texture), some geometric descriptor, motion field, etc. Conversely, learnable spectral descriptors try to learn the best spectral kernel that acts on the Laplacian eigenvalues, thus limited to geometric data of the manifold only.

\paragraph*{Geodesic CNN (GCNN)}
was introduced by Masci \etal \cite{masci2015shapenet} as a generalization of CNN to triangular meshes based on geodesic local patches. 
The core of this method is the construction of local geodesic polar coordinates 
using a procedure previously employed for intrinsic shape context descriptors \cite{ISC}. The {\em patch operator} $(D(x)f)(\theta,\rho)$ in GCNN maps the values of the function $f$ around vertex $x$ into the local polar coordinates $\theta, \rho$, leading to the definition of the {\em geodesic convolution} 
\begin{eqnarray}
\label{eq:geoconv}
(f \ast a)(x) &=& \max_{\Delta\theta \in [0, 2\pi)} \int  a(\theta + \Delta\theta,\rho) (D(x)f)(\theta,\rho) d\rho d\theta, 
\end{eqnarray}
which follows the idea of multiplication by template, but is defined up to arbitrary rotation $\Delta\theta \in [0, 2\pi)$ due to the ambiguity in the selection of the origin of the angular coordinate. 
Taking the maximum over all possible rotations of the template $a(\rho,\theta)$ is necessary to remove this ambiguity. 
Here, and in the following, $f$ is some feature vector that is defined on the surface (e.g. texture, geometric descriptors, etc.)

There are several drawbacks to this construction. First, the charting method relies on a fast marching-like procedure requiring a triangular mesh. The method is relatively insensitive to the triangulation \cite{ISC}, but may fail if the mesh is very irregular. 
Second, the radius of the geodesic patches must be sufficiently small compared to the injectivity radius of the shape, otherwise the resulting patch is not guaranteed to be a topological disk. In practice, this limits the size of the patches one can safely use, or requires an adaptive radius selection mechanism.

\paragraph*{Spectral CNN (SCNN). }
Bruna \etal \cite{Bruna} defined a generalization of convolution in the spectral domain, appealing to the Convolution Theorem, stating that in the Euclidean case, the convolution operator is diagonalized in the Fourier basis. This allows defining a non-shift-invariant convolution as the inverse Fourier transform of the product of two Fourier transforms, 
\begin{eqnarray}
\label{eq:geoconv}
(f \ast a)(x) &=&  \sum_{k\geq 0} \langle f, \phi_k \rangle_X \underbrace{\langle a, \phi_k \rangle_X}_{\hat{a}_k} \phi_k(x),
\end{eqnarray}
where the Fourier transform is understood as inner products of the function with the Laplace-Beltrami orthogonal eigenfunctions. 
The filter in this formulation is represented in the frequency domain, by the set of Fourier coefficients $\{\hat{a}_k\}_{k\geq 0}$. The SCNN is essentially a classical CNN where standard convolutions are replaced by definition~(\ref{eq:geoconv}) and the frequency representations of the filters are learned. 

The key drawback of this approach is the lack of generalizability, due to the fact that the filter coefficients $\{\hat{a}_k\}_{k\geq 0}$ depend on the basis $\{\phi_k(x)\}_{k\geq 0}$; as a result, applying a filter on two different shapes may produce two different results. 
Therefore, SCNN can be used to learn on a non-Euclidean domain, but not across different domains. 
Secondly, the filters defined in the frequency domain lack a clear geometric interpretation. Third, the filters are not guaranteed to be localized in the spatial domain.

\paragraph*{Localized Spectral CNN (LSCNN).}
Boscaini \etal \cite{WFT2015} proposed an approach that combines the ideas of GCNN (spatial ``patch operator'') and SCNN (frequency-domain filters). The key concept is based on the Windowed Fourier Transform (WFT) \cite{Shumann}, a generalization to manifolds of a classical tool of signal processing suggesting to apply frequency analysis in a small window. 
The WFT boils down to projecting the function $f$ on a set of modulated local windows (``atoms'') 
$g_{\xi,k}(x) = \phi_k(x) \sum_{l\geq 0} \hat{g}_l \phi_l(\xi) \phi_l(x)$, 
\begin{eqnarray}
\label{eq:wft}
(S(x) f)_k &=&  \langle f, g_{x,k}\rangle_X  = \sum_{l\geq 0} \hat{g}_l \phi_l(x) \langle f, \phi_l \phi_k \rangle_X, 
\end{eqnarray}
where $\{\hat{g}_k\}_{k\geq 0}$ are the Fourier coefficients of the window. 
The WFT can be regarded as a ``patch operator'', representing the local values of $f$ around a point $x$ in the frequency domain. 

The main advantage of this approach is that being a spectral construction, it is easily applicable to any representations of the shape (mesh, point cloud, etc.), provided an appropriate discretization of the Laplace-Beltrami operator. 
%
%
Since the patch operator is constructed in the frequency domain using the WFT, there is also no issue related to the topology of the patch like in the GCNN. 
Yet, unlike the geodesic patches used in GCNN, the disadvantage of LSCNN is the lack of oriented structures which tend to be important in capturing the local context. 
From the computational standpoint, a notable disadvantage of the WFT-based construction is the need to explicitly produce each window, which result in high memory and computational requirements.

\section{Anisotropic convolutional neural networks}
\label{sec:our}

The construction presented in this paper aims at benefiting from all the advantages of the aforementioned intrinsic CNN approaches, without inheriting their drawbacks.

\paragraph*{Intrinsic convolution.}
The key idea of the Anisotropic CNN presented in this paper is the construction of a patch operator using anisotropic heat kernels. We interpret heat kernels as local weighting functions and construct 
\begin{equation}
    \label{eq:acnn}
    (D(x) f)(\theta, t) =\frac{ \int_X h_{\alpha\theta t}(x,\xi) f(\xi) d\xi }{ \int_X h_{\alpha\theta t}(x,\xi) d\xi}, 
\end{equation}
for some anisotropy level $\alpha > 1$. This way, the values of $f$ around point $x$ are mapped to a local system of coordinates $(\theta, t)$ that behaves like a polar system (here $t$ denotes the scale of the heat kernel and $\theta$ is its orientation). 

We define {\em intrinsic convolution} as 
\begin{eqnarray}
\label{eq:intconv}
(f \ast a)(x) &=& \int  a(\theta,t) (D(x)f)(\theta,t) dt d\theta, 
\end{eqnarray}
Note that unlike the arbitrarily oriented geodesic patches in GCNN, necessitating to take a maximum over all the template rotations~(\ref{eq:geoconv}), in our construction it is natural to use the principal curvature direction as the reference $\theta = 0$. 

Such an approach has a few major advantages compared to previous intrinsic CNN models. 
First, being a spectral construction, our patch operator can be applied to any shape representation (like LSCNN and unlike GCNN). 
Second, being defined in the spatial domain, the patches and the resulting filters have a clear geometric interpretation (unlike LSCNN). 
Third, our construction accounts for local directional patterns (like GCNN and unlike LSCNN).
Fourth, the heat kernels are always well defined independently of the injectivity radius of the manifold (unlike GCNN).  
We summarize the comparative advantages in Table~\ref{tab:cnns}.

\begin{table*}
\centering
\begin{tabular}{ l c c c c c c c}
Method &  Representation & Input & Generalizable & Filters & Context & Directional & Task\\
\hline
 OSD \cite{OSD} & Any & Geometry & Yes & Spectral & No & No & Descriptor\\
 ADD \cite{ADD} & Any & Geometry & Yes & Spectral & No & Yes & Any\\
 RF \cite{rodoladense} & Any & Any & Yes & Spectral & No & No & Correspondence\\ 
  GCNN \cite{masci2015shapenet} & Mesh & Any  & Yes & Spatial & Yes & Yes & Any\\
  SCNN \cite{Bruna} & Any & Any  & No & Spectral & Yes & No & Any\\
  LSCNN \cite{WFT2015} & Any & Any  & Yes & Spectral & Yes & No & Any\\
  Proposed ACNN & Any & Any  & Yes & Spatial & Yes & Yes & Any\\
  \hline
\end{tabular}
\caption{\label{tab:cnns}Comparison of different intrinsic learning models. Our ACNN model combines all the best properties of the other models. 
Note that OSD and ADD are local spectral descriptors operating with intrinsic geometric information of the shape and cannot be applied to arbitrary input, unlike the Random Forest (RF) and convolutional models. 
}
\end{table*}

\subsection{ACNN architecture}


Similarly to Euclidean CNNs, our ACNN consists of several layers that are applied subsequently, i.e. the output of the previous layer is used as the input into the subsequent one. The network is called {\em deep} if many layers are employed.  
ACNN is applied in a point-wise manner on a function defined on the manifolds, producing a point-wise output that is interpreted as soft correspondence, as described below. 
We distinguish between the following types of layers:

\paragraph*{Fully connected (FC$Q$)} layer
typically follows the input layer and precedes the output layer to adjust the 
input and output dimensions by 
means of a linear combination. Given a $P$-dimensional input $\mathbf{f}^\mathrm{in}(x) = ( f_1^\mathrm{in}(x), \hdots, f_P^\mathrm{in}(x))$, the FC layer produces a $Q$-dimensional output $\mathbf{f}^\mathrm{out}(x) = ( f_1^\mathrm{out}(x), \hdots, f_Q^\mathrm{out}(x))$ as a linear combination of the input components with learnable weights $w$, 
\begin{equation}
f^\text{out}_{q}(x) =
\eta\left( \sum_{p=1}^P w_{qp} f^\text{in}_p(x) \right); 
\quad q = 1,\hdots, Q, 
\label{eq:cnn_fc}
\end{equation}
The output of the FC layer is optionally passed through a non-linear function such as the ReLU\cite{nair2010}, $\eta(t) = \max\{0,t\}$.

\paragraph*{Intrinsic convolution (IC$Q$)} layer 
replaces the convolutional layer used in classical Euclidean CNNs with the construction~(\ref{eq:intconv}).  
The IC layer contains $PQ$ filters arranged in banks ($P$ filters in $Q$ banks); each bank corresponds to an output dimension. The filters are applied to the input as follows, 
\begin{equation}
\label{eq:gcnn}
f^\mathrm{out}_{q}(x) = 
\sum_{p=1}^P (f^\text{in}_p \star a_{qp})(x),
\quad q = 1,\hdots, Q, 
\end{equation}
where $a_{qp}(\theta,t)$ are the learnable coefficients of the $p$th filter in the $q$th filter bank.

\paragraph*{Softmax} layer is used as the output layer in a particular architecture employed for learning correspondence; it applies the softmax function 
\begin{equation}
\label{eq:softmax}
f^\mathrm{out}_{p}(x) = 
\sigma(f^\mathrm{in}_{p}(x)) = 
\frac{e^{f^\mathrm{in}_{p}(x)}}{\sum_{p=1}^P e^{f^\mathrm{in}_{p}(x)}} 
\end{equation}
to the $P$-dimensional input. The result is a vector that can be interpreted as a probability distribution.

\paragraph*{Dropout($\pi$)} layer~\cite{hinton2012dropout} is a fixed layer that injects binomial noise to each of the computational units of the network; it has been shown to be an excellent remedy to prevent overfitting.
During training, an i.i.d. binary mask $m_p \sim \mathrm{Binomial}(\pi_\mathrm{drop})$ is generated for each input dimension; each element is $1$ with probability $1 - \pi_\mathrm{drop}$, 
\begin{equation}
\label{eq:dropout_train}
f^\mathrm{out}_{p}(x) = 
m_p  f^\mathrm{in}_{p}(x). 
\end{equation}
%
At test time, in order to do inference one would have to integrate over all possible binary masks. 
However, it has been shown that rescaling the input by the drop probability of the layer, 
\begin{equation}
\label{eq:dropout_train}
f^\mathrm{out}_{p}(x) = 
\pi_\mathrm{drop}  f^\mathrm{in}_{p}(x). 
\end{equation}
 is a good approximation  applicable for real applications.

\paragraph*{Batch normalization} layer is another fixed layer recently introduced in~\cite{ioffe2015} to reduce training times of very large CNN models. 
It normalizes each mini-batch during stochastic optimization to have zero mean and unit variance, and then performs a linear transformation of the form
\begin{equation}
\label{eq:batch_norm}
f^\mathrm{out}_{p}(x) = \frac{f^\mathrm{in}_{p}(x) - \mu}{\sqrt{\sigma^2 + \epsilon}} \gamma + \beta
\end{equation}
where $\mu$ and $\sigma^2$ are, respectively, the mean and the variance of the data estimated on the training set using exponential moving average; $\epsilon$ is a small positive constant to avoid numerical errors.
After training, one can re-estimate the statistics on the test set or simply keep the training set estimates.

Overall, the ACNN architecture combining several layers of different type, acts as a non-linear parametric mapping of the form $\mathbf{f}_\Theta(x)$ at each point $x$ of the shape, where $\Theta$ denotes the set of all learnable parameters of the network. 
The choice of the parameters is done by an optimization process, minimizing a task-specific cost. 
Here, we focus on learning shape correspondence.


\subsection{Learning correspondence}


Finding the correspondence in a collection of shapes can be posed as a labelling problem, where one tries to label each vertex of a given {\em query} shape $X$ with the index of a corresponding point on some common {\em reference} shape $Y$  \cite{rodoladense}. Let $n$ and $m$ denote the number of vertices in $X$ and $Y$, respectively. 
For a point $x$ on a query shape, the output of ACNN $\mathbf{f}_\Theta(x)$ is $m$-dimensional and is interpreted as a probability distribution (`soft correspondence') on $Y$. 
The output of the network at all the points of the query shape can be arranged as an $n\times m$ matrix with elements of the form $f_\Theta(x,y)$, representing the probability of $x$ mapped to $y$.

Let us denote by $y^*(x)$ the ground-truth correspondence of $x$ on the reference shape. We assume to be provided with examples of points from shapes across the collection and their ground-truth correspondence, $\mathcal{T} = \{ (x, y^*(x)) \}$. 
 The optimal parameters of the network are found by minimizing the {\em multinomial regression loss}
\begin{eqnarray}
    \label{eq:corresp}
    \ell_{\mathrm{reg}}(\boldsymbol{\Theta}) 
    &=& -\sum_{(x, y^*(x))\in \mathcal{T}} \log \boldsymbol{f}_{\Theta}(x,y^*(x)),
\end{eqnarray}
which represents the Kullback-Leibler divergence between the probability distribution produced by the network and the groundtruth distribution $\delta_{y^*(x)}$.

\subsection{Correspondence refinement}

\paragraph*{Full correspondence. } 
The most straightforward way to convert the soft correspondence $f(x,y)$ produced by ACNN into a point-wise correspondence is by assigning $x$ to 
\begin{equation}
\hat{y}(x) = \mathrm{arg}\max_{y\in Y} f(x,y). 
\end{equation}
The value $c(x) = \max_{y \in Y} f(x,y) \in [0,1]$ can be interpreted as the {\em confidence} of the prediction: the closer the distribution produced by the network is to a delta-function (in which case $c=1$), the better it is.

In this paper, we use a slightly more elaborate scheme to refine the soft correspondences produced by ACNN. First, we select a subset of points $I = \{x : c(x) > \tau_\mathrm{th}\}$ at which the confidence of the  predicted correspondence exceeds some threshold $\tau_\mathrm{th}$. 
%
Second, we use this subset of corresponding points 
to find a functional map \cite{ovsjanikov2012functional} between $L^2(X)$ and $L^2(Y)$ by solving the linear system of $|I|k$ equations in $k^2$ variables, 
\begin{eqnarray}
    \label{eq:fm}
   \pmb{\Phi}_I\mathbf{C} &=& \pmb{\Psi}_I, 
\end{eqnarray}
where 
\begin{eqnarray}
   \pmb{\Phi}_I &=& (\phi_1(x), \hdots, \phi_k(x)) \,\,\,\,\, x\in I,  \\
      \pmb{\Psi}_I &=& (\psi_1(\hat{y}(x)), \hdots, \psi_k(\hat{y}(x))) \,\,\,\,\, x\in I, 
\end{eqnarray}
are the first $k$ Laplace-Beltrami eigenfunctions of shapes $X$ and $Y$, respectively, sampled at the subset of corresponding points (represented as $|I| \times k$ matrices). 
The $k\times k$ matrix $\mathbf{C}$ represents the functional correspondence between $L^2(X)$ and $L^2(Y)$ in the frequency domain. 
The parameters $\tau_\mathrm{th}$ and $k$ must be chosen in such a way that the system is over-determined, i.e. $|I| > k$. 
Third, after having found $\mathbf{C}^*$ by solving~(\ref{eq:fm}) in the least-squares sense, we produce a new point-wise correspondence by matching $\pmb{\Phi}\mathbf{C}^*$ and $\pmb{\Psi}$ in the $k$-dimensional eigenspace,  
\begin{equation}
y(x) = \mathrm{arg}\max_{y\in Y} \|  (\phi_1(x), \hdots, \phi_k(x))\mathbf{C}^* - (\psi_1(y), \hdots, \psi_k(y) \|_2. 
\end{equation}

\paragraph*{Partial correspondence. }
 A similar procedure is employed in the setting of partial correspondence, where instead of the computation of a functional map, we use the recently introduced {\em partial functional map} \cite{PFM2016}.


\section{Numerical implementation}
\label{sec:discrete}

\definecolor{mygray}{HTML}{6195C8}

\begin{figure}[t!]
  \centering
	\begin{overpic}
		[trim=0cm 0cm 0cm 0cm,width=0.97\linewidth]{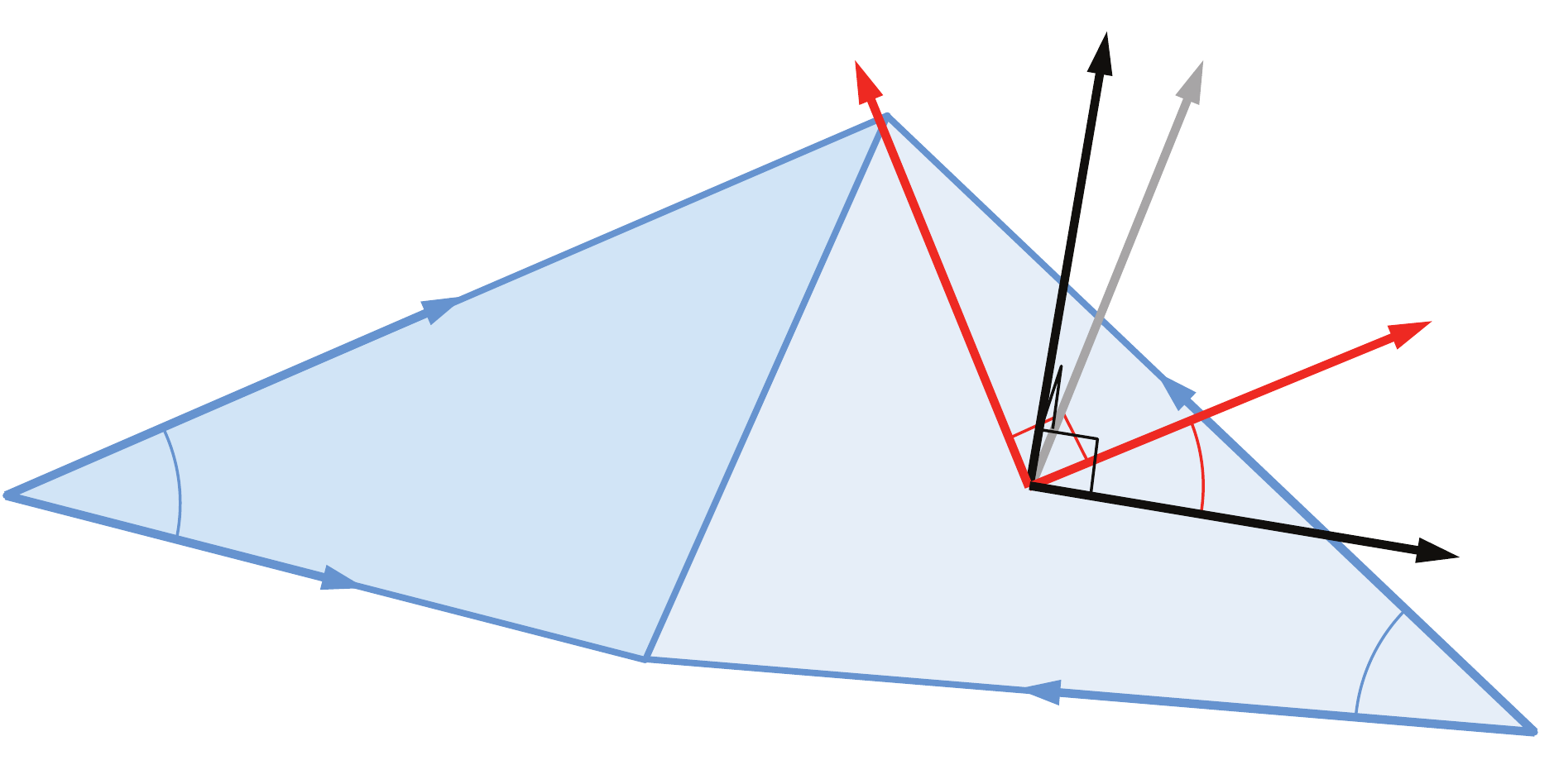}
		\put(89,6){\footnotesize $\alpha_{ij}$}
		\put(6,18){\footnotesize $\beta_{ij}$}
		\put(73,19){\footnotesize $\color{red}\theta$}
		\put(40.2,5){\footnotesize $i$}
		\put(58,44){\footnotesize $j$}
		\put(100,2){\footnotesize $k$}
		\put(-2,18){\footnotesize $h$}
		\put(52,48){\footnotesize $\color{red}\mathbf{R}_\theta\hat{\vct{u}}_m$}
		\put(93,29){\footnotesize $\color{red}\mathbf{R}_\theta\hat{\vct{u}}_M$}
		\put(70,49){\footnotesize $\hat{\vct{u}}_m$}
		\put(94.5,14){\footnotesize $\hat{\vct{u}}_M$}
		\put(78,47){\footnotesize $\color{gray}\hat{\vct{n}}$}
		\put(75,28){\footnotesize $\color{mygray}\hat{\vct{e}}_{kj}$}
		\put(65,2.5){\footnotesize $\color{mygray}\hat{\vct{e}}_{ki}$}
		\put(21,9){\footnotesize $\color{mygray}\hat{\vct{e}}_{hi}$}
		\put(25.5,33){\footnotesize $\color{mygray}\hat{\vct{e}}_{hj}$}
	\end{overpic}
	\caption{\label{fig:cotans}Discretization of the anisotropic Laplace-Beltrami operators on a triangular mesh. The orthogonal basis vectors $\hat{\vct{u}}_M,\hat{\vct{u}}_m$, as well as their rotated counterparts (in red), lie in the plane of the respective triangle (reproduced from \cite{ADD}). 
	}
\end{figure}

\subsection{Anisotropic Laplacian discretization}

%
In the discrete setting, the surface $X$ is sampled at $n$ points $V = \{ \mathbf{x}_1,\dots, \mathbf{x}_n \}$. The points are connected by edges $E$ and faces $F$, forming a manifold triangular mesh $(V,E,F)$. 
%
%
To each triangle $ijk \in F$, we attach an orthonormal reference frame $\vct{U}_{ijk} = (\hat{\vct{u}}_M,\hat{\vct{u}}_m,\hat{\vct{n}})$, where $\hat{\vct{n}}$ is the unit normal vector to the triangle and $\hat{\vct{u}}_M,\hat{\vct{u}}_m\in\mathbb{R}^3$ are the directions of principal curvature, computed using the method of \cite{cohen03}. 
%
%
The thermal conductivity tensor for the triangle $ijk$ operating on tangent vectors is expressed w.r.t. $\vct{U}_{ijk}$ as a  $3\times 3$ matrix $\begin{psmallmatrix}\alpha& &\\ & 1 & \\ & &1\end{psmallmatrix}$.
We first derive the case $\theta = 0$. 

Let $\hat{\vct{e}}_{ab} \in \mathbb{R}^3$ denote the oriented edge pointing from vertex $a$ to vertex $b$, normalized to unit length, and consider the triangle $ijk$ as in Figure~\ref{fig:cotans}. We define the $\mathbf{H}$-weighted inner product between edges $\hat{\vct{e}}_{kj}$ and $\hat{\vct{e}}_{ki}$ as
\begin{align}
\label{eq:shear}
\langle \hat{\vct{e}}_{kj} , \hat{\vct{e}}_{ki}\rangle_{\mathbf{H}} &
=\hat{\vct{e}}_{kj}\T \underbrace{\vct{U}_{ijk} \begin{psmallmatrix}\alpha& &\\ & 1 & \\ & &1\end{psmallmatrix} \vct{U}_{ijk}\T}_\vct{H} \hat{\vct{e}}_{ki}\,,
\end{align}
%
where the {\em shear matrix} $\vct{H}$ encodes the anisotropic scaling up to an orthogonal basis change. Note that in the isotropic case ($\alpha = 1$) we have $\vct{H}=\mathbf{I}$, such that the $\mathbf{H}$-weighted inner product simplifies to the standard inner product $\langle \hat{\vct{e}}_{kj} , \hat{\vct{e}}_{ki}\rangle_{\mathbf{H}} = \cos \alpha_{ij}$.

The discretization of the anisotropic Laplacian takes the form of an $n \times n$ sparse matrix $\vct{L} = -\vct{S}^{-1} \vct{W}$. 
The {\em mass matrix}  $\vct{S}$ is a diagonal matrix of area elements $s_i = \frac{1}{3}\sum_{jk:ijk\in F} A_{ijk}$, where $A_{ijk}$ denotes the area of triangle $ijk$. The {\em stiffness matrix} $\vct{W}$ is composed of weights 
\begin{eqnarray}\label{eq:acotan}
w_{ij} & = & \left\{ 
		\begin{array}{ll}
			\frac{1}{2}\left(\frac{\langle \hat{\vct{e}}_{kj} , \hat{\vct{e}}_{ki}\rangle_{\mathbf{H}}}{\sin \alpha_{ij}} + \frac{\langle \hat{\vct{e}}_{hj} , \hat{\vct{e}}_{hi}\rangle_{\mathbf{H}}}{\sin \beta_{ij}}\right) &   (i,j) \in E_\mathrm{int}; \\ 
			\frac{1}{2}\frac{\langle \hat{\vct{e}}_{kj} , \hat{\vct{e}}_{ki}\rangle_{\mathbf{H}}}{\sin \alpha_{ij}}  &   (i,j) \in E_{\partial}; \\ 
			-\sum_{k\neq i} w_{ik}    & i = j; \\
			0 & \mathrm{else}\,,
		\end{array}
\right.
\end{eqnarray}
where the notation is according to Figure~\ref{fig:cotans} and $E_\mathrm{int}, E_{\partial}$ denote interior and boundary edges, respectively. 
%
In the isotropic case, 
$\frac{\langle \hat{\vct{e}}_{kj},\hat{\vct{e}}_{ki}\rangle_{\mathbf{H}}}{\sin \alpha_{ij}} = \frac{\cos \alpha_{ij}}{\sin \alpha_{ij}}=\cot \alpha_{ij}$, thus reducing equation~\eqref{eq:acotan} to the classical cotangent formula \cite{macneal1949solution,duffin1959distributed,Pinkall1993,meyer2003:ddg}.


To obtain the general case $\theta \neq 0$, it is sufficient to rotate the basis vectors $\vct{U}_{ijk}$ on each triangle around the respective normal $\hat{\vct{n}}$ by the angle $\theta$, equal for all triangles (see Figure~\ref{fig:cotans}, red). Denoting by $\vct{R}_\theta$ the corresponding $3\times 3$ rotation matrix, this is equivalent to modifying the $\mathbf{H}$-weighted inner product with the directed shear matrix $\vct{H}_\theta = \vct{R}_\theta \vct{H}\vct{R}\T_\theta$. The resulting weights $w_{ij}$ in equation \eqref{eq:acotan} are thus obtained by using the inner products $\langle \hat{\vct{e}}_{kj} , \hat{\vct{e}}_{ki}\rangle_{\mathbf{H}_\theta} = \hat{\vct{e}}_{kj}\T \vct{H}_\theta \hat{\vct{e}}_{ki}$.

\subsection{Heat kernels}
The computation of heat kernels is performed in the frequency domain, using a truncation of formula~(\ref{eq:heat_kernel}). 
The first $k$ eigenfunctions and eigenvalues of the Laplacian are computed by performing the generalized eigen-decomposition  $\mathbf{W} \pmb{\Phi} = \mathbf{S}\pmb{\Phi}\pmb{\Lambda}$, where $\pmb{\Phi} = (\pmb{\phi}_1, \hdots, \pmb{\phi}_k)$ is an $n\times k$ matrix containing as columns the discretized eigenfunctions and $\pmb{\Lambda} = \mathrm{diag}(\lambda_1, \hdots, \lambda_k)$ is the diagonal matrix of the corresponding eigenvalues. 
%
The heat operator is given by $\pmb{\Phi} e^{-t\pmb{\Lambda}} \pmb{\Phi}^\top$; the $i$th row/column represents the values of the heat kernel at point $\mathbf{x}_i$.

\section{Results}
\label{sec:results}

In this section, we evaluate the proposed ACNN method and compare it to state-of-the-art approaches on the FAUST  \cite{FAUST} and SHREC'16 Partial Correspondence \cite{SHREC2016p} benchmarks. 

\subsection{Settings}

%

\paragraph*{Implementation.} 
Isotropic Laplacians were computed using the cotangent formula \cite{macneal1949solution,duffin1959distributed,Pinkall1993,meyer2003:ddg}; anisotropic Laplacians were computed according to~(\ref{eq:acotan}). 
Heat kernels were computed in the frequency domain using all the eigenpairs. 
In all experiments, we used $L=16$ orientations and the anisotropy parameter $\alpha = 100$. 

Neural networks were implemented in Theano~\cite{bergstra2010theano}. 
The ADAM~\cite{kingma2014} stochastic optimization algorithm was used with 
initial learning rate of $10^{-3}$, $\beta_1=0.9$, and $\beta_2=0.999$. 
As the input to the networks, we used the local SHOT descriptor \cite{SHOT} with $544$ dimensions and using default parameters. 
For all experiments, training was done by minimizing the loss~(\ref{eq:corresp}). 
The code to reproduce all the experiments in this paper, and the full framework will be released upon publication.


\paragraph*{Timing.} 
The following are typical timings for FAUST shapes with 6.9K vertices. Laplacian computation and eigendecomposition took $1$ sec 
and $4$ seconds per angle, respectively on a desktop workstation with 64Gb of RAM and i7-4820K CPU.
Forward propagation of the trained model takes approximately $0.5$ sec to produce the dense soft correspondence for all the vertices.

\begin{figure}[t!]
\begin{overpic}
[width=1\linewidth]{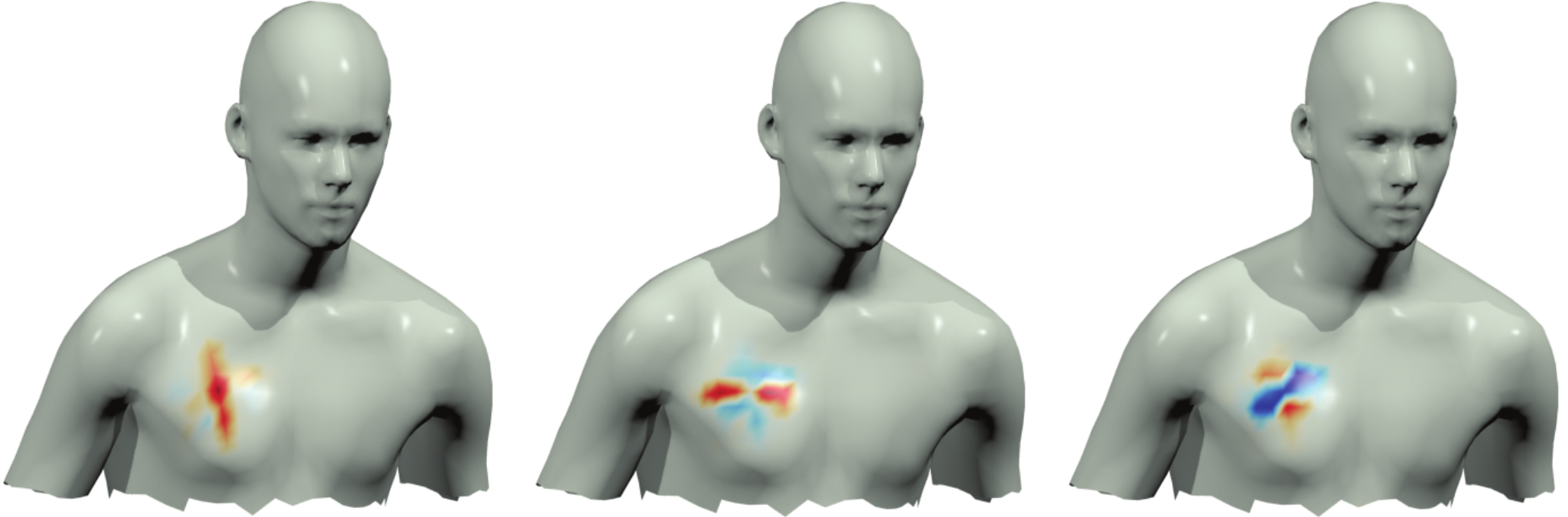}
 \end{overpic}
  \centering
   \caption{ \label{fig:filters} Examples of filters in the first IC layer learned by the ACNN (hot and cold colors represent positive and negative values, respectively). 
   }
   \vspace{-3mm}
\end{figure}

\subsection{Full mesh correspondence}
\paragraph*{Data.}
In the first experiment, we used the FAUST humans dataset \cite{FAUST}, containing $100$ meshes of $10$ scanned subjects, each in $10$ different poses. The shapes in the collection manifest strong non-isometric deformations. Vertex-wise groundtruth correspondence is known between all the shapes.
The zeroth FAUST shape containing $6890$ vertices was used as reference; for each
point on the query shape, the output of the network represents the soft
correspondence as a $6890$-dimensional vector which was then converted to point correspondence 
with the technique explained in Section~\ref{sec:our}.  First $80$ shapes for training and the remaining $20$ for testing, following verbatim the settings of~\cite{masci2015shapenet}. 
Batch normalization~\cite{ioffe2015} was used to speed up the training; we did not experience any noticeable difference in raw performance of the produced soft correspondence compared to un-normalized setting.

\begin{figure*}[t!]
\begin{overpic}
[width=0.975\linewidth]{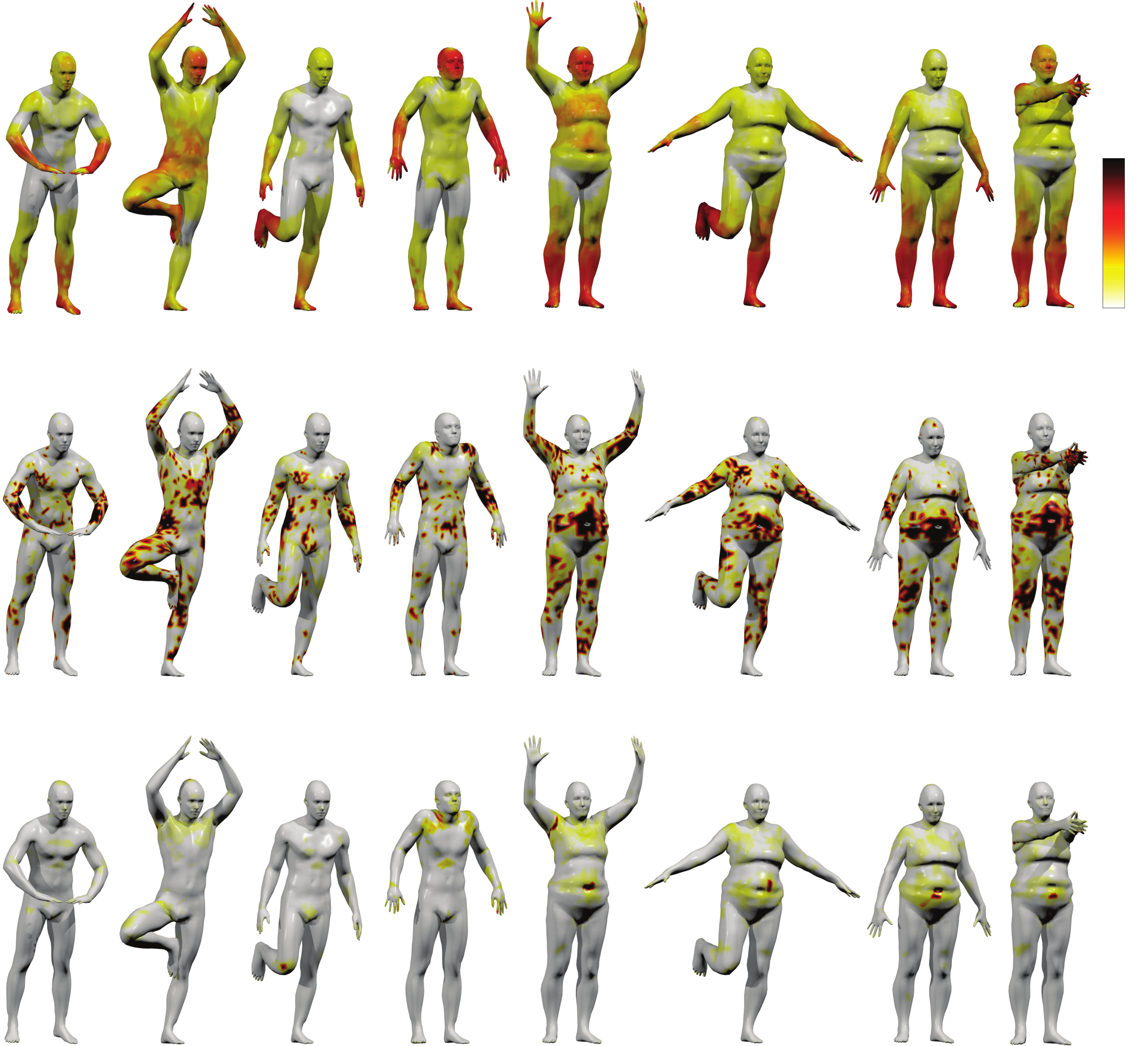}
  \put(45,-2){\footnotesize Anisotropic CNN}
 \put(44.5,29.1){\footnotesize Geodesic CNN}
 \put(42,62){\footnotesize Blended Intrinsic Map}
 \put(98.3,64){\tiny 0} 
 \put(98.3,77){\tiny 0.1} 
 \end{overpic}
  \centering
  \vspace{3mm}
   \caption{ \label{fig:faust_geoerr} Pointwise geodesic error (in $\%$ of geodesic diameter) of different correspondence methods (top to bottom: Blended intrinsic maps, GCNN, ACNN) on the FAUST dataset. For visualization clarity, the error values are saturated at $10\%$ of the geodesic diameter. Hot colors correspond to large errors. 
   Note the different behavior of different approaches: BIM produces large distortions with very few accurate matches; GCNN produces many near-perfect matches but also many matches with large distortion; ACNN produces very few matches with large distortion and many near-perfect matches. 
   }
\end{figure*}

\paragraph*{Methods.}
Batch normalization allows us to effectively train larger and deeper networks, for this experiment
we adopted the following architecture inspired by GCNN~\cite{masci2015shapenet}: FC$64$+IC$64$+IC$128$+IC$256$+FC$1024$+FC$512$+Softmax. 
%
%
Additionally, we compared our method to Random Forests (RF) \cite{rodoladense}, Blended Intrinsic Maps (BIM) \cite{kim2011blended}, Localized Spectral CNN (LSCNN) \cite{WFT2015}, and Anisotropic Diffusion Descriptors (ADD) \cite{ADD}.

\begin{figure}[t!]
\setlength\figureheight{5cm} 
\setlength\figurewidth{\linewidth}
\input{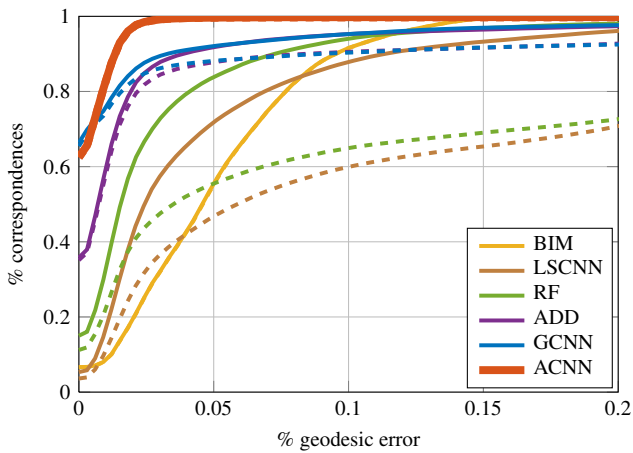}
\caption{\label{fig:faust_plot} Performance of different correspondence methods on FAUST meshes.  Evaluation of the correspondence was done using the symmetric (solid) and asymmetric (dashed) Princeton protocol. }
\end{figure}

\paragraph*{Results.}
Figure~\ref{fig:faust_plot} shows the performance of different methods. The performance was evaluated using the Princeton protocol \cite{kim2011blended}, plotting
the percentage of matches that are at most $r$-geodesically distant from the groundtruth correspondence on the reference shape. Two versions of the protocol consider intrinsically symmetric matches as correct (symmetric setting) or wrong (asymmetric, more challenging setting). Some methods based on intrinsic structures (e.g. LSCNN or RF applied on WKS descriptors) are invariant under intrinsic symmetries and thus cannot distinguish between symmetric points.  
The proposed ACNN method clearly outperforms all the compared approaches and also perfectly distinguishes symmetric points.

Figure~\ref{fig:teaser} visualizes some correspondences obtained by ACNN using texture mapping. The correspondence show almost no noticeable artifacts. 
Figure~\ref{fig:faust_geoerr} shows the pointwise geodesic error of different correspondence methods (distance of the correspondence at a point from the groundtruth). ACNN shows dramatically smaller distortions compared to other methods. Over $60\%$ of matches are exact (zero geodesic error), while only a few points have geodesic error larger than $10\%$ of the geodesic diameter of the shape.

\subsection{Partial correspondence}
\paragraph*{Data.}
In the second experiment, we used the recent very challenging SHREC'16 Partial Correspondence benchmark \cite{SHREC2016p}, consisting of nearly-isometrically deformed shapes from eight classes, with different parts removed. Two types of partiality in the benchmark are {\em cuts} (removal of a few large parts) and {\em holes} (removal of many small parts). In each class, the vertex-wise groundtruth correspondence between the full shape and its partial versions is given. 
The dataset was split into training and testing disjoint sets. For cuts, training was done on 15 shapes per class; for holes, training was done on 10 shapes per class. 
%

\paragraph*{Methods.}
We used the following ACNN architecture: IC$32$+FC$1024$+DO($0.5$)+FC$2048$+DO($0.5$)+Softmax. The dropout regularization, with $\pi_\mathrm{drop} = 0.5$, was crucial to avoid overfitting on such a small training set. 
We compared ACNN to RF \cite{rodoladense} and Partial Functional Maps (PFM) \cite{PFM2016}. 
For the evaluation, we used the protocol of \cite{SHREC2016p}, which closely follows the Princeton benchmark.

\begin{figure*}[t!]
\vspace{-1mm}
\begin{overpic}
 [width=1\linewidth]{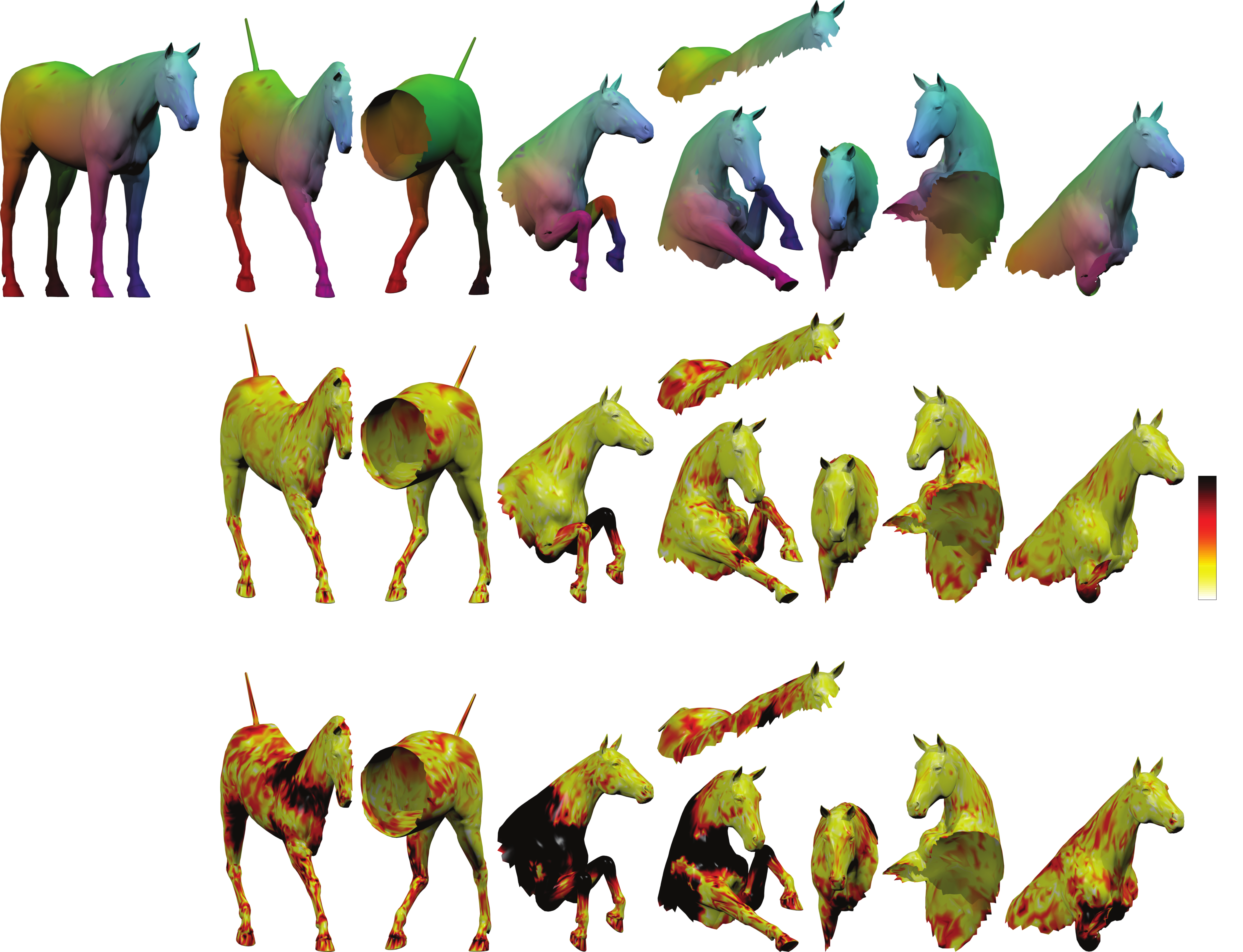}
 \put(50,-1){\footnotesize Random Forest}
 \put(49,26){\footnotesize Anisotropic CNN}
 \put(98.1,28){\tiny 0} 
 \put(98.1,38){\tiny 0.1} 
 \end{overpic}
  \centering
  \vspace{1.5mm}
   \caption{ \label{fig:horse_partial}Examples of partial correspondence on the horse shape from the SHREC'16 Partial (cuts) dataset. First row: correspondence produced by ACNN. Corresponding points are shown in similar color. Reference shape is shown on the left. 
   Second and third rows: pointwise geodesic error (in $\%$ of geodesic diameter) of the ACNN and RF correspondence, respectively. For visualization clarity, the error values are saturated at $10\%$ of the geodesic diameter. Hot colors correspond to large errors.}
\end{figure*}

\paragraph*{Cuts.}
Figure~\ref{fig:shrec_cuts} compares the performance of different partial matching methods on the SHREC'16 Partial (cuts) dataset. ACNN outperforms other approaches with a significant margin. 
Figure~\ref{fig:horse_partial} (top) shows examples of partial correspondence on the horse shape as well as the pointwise geodesic error (bottom). We observe that the proposed approach produces high-quality correspondences even in such a challenging setting. 

\begin{figure}[t!]
\setlength\figureheight{5cm} 
\setlength\figurewidth{\linewidth}
\input{./kim_cuts.tikz}
\caption{\label{fig:shrec_cuts}Performance of different correspondence methods on SHREC'16 Partial (cuts) meshes. Evaluation of the correspondence was done using the symmetric Princeton protocol. }
\end{figure}

\begin{figure}[t!]
\setlength\figureheight{5cm} 
\setlength\figurewidth{\linewidth}
\input{./kim_holes.tikz}
\caption{\label{fig:shrec_holes}Performance of different correspondence methods on SHREC'16 Partial (holes) meshes. Evaluation of the correspondence was done using the symmetric Princeton protocol. }
\end{figure}

\begin{figure*}[t!]
\vspace{-1mm}
\begin{overpic}
 [width=1\linewidth]{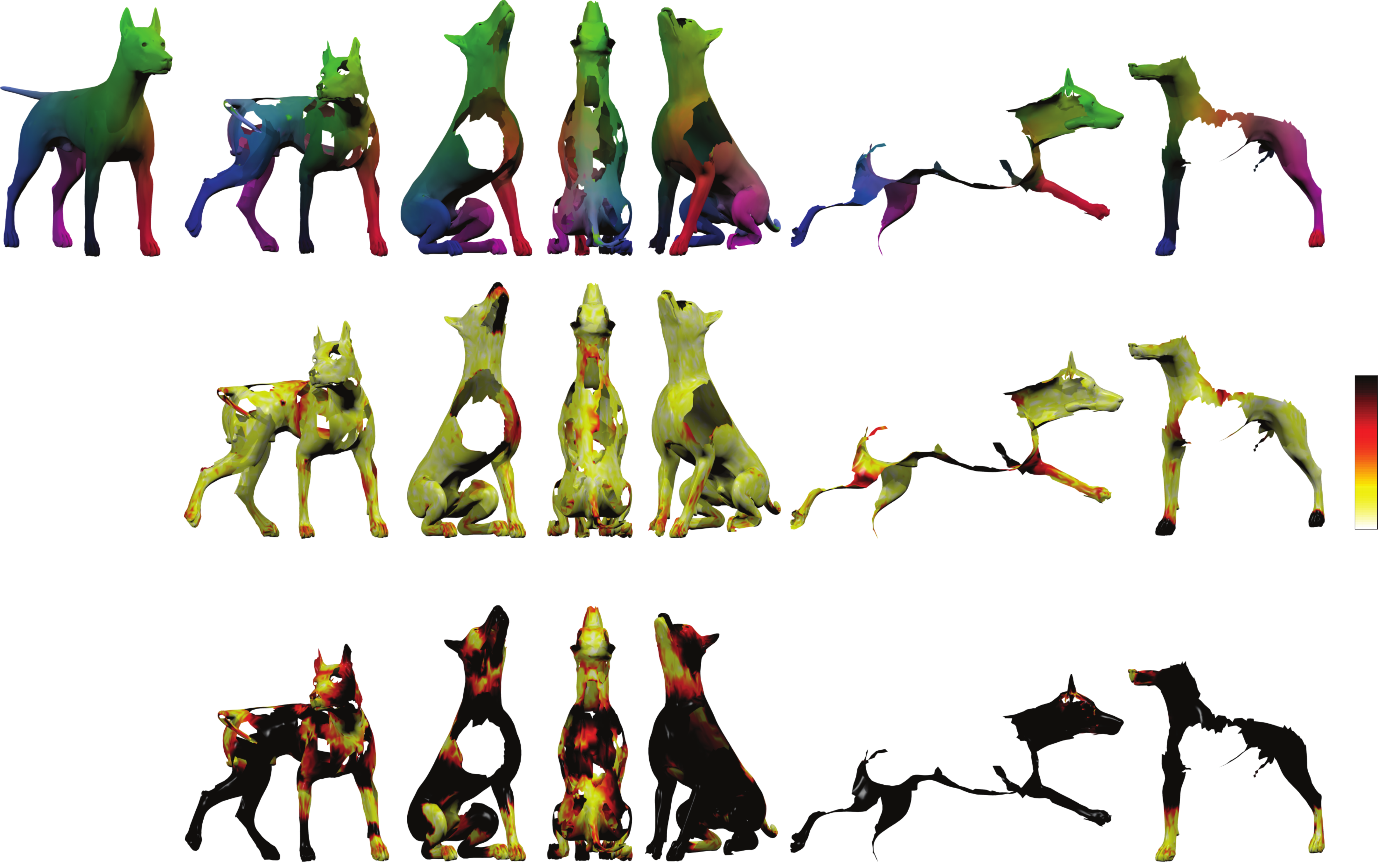}
 \put(50,-2.25){\footnotesize Random Forest}
 \put(49,20.5){\footnotesize Anisotropic CNN}
 \put(98.8,23.75){\tiny 0} 
 \put(98.8,34){\tiny 0.1} 
 \end{overpic}
  \centering
  \vspace{1.5mm}
   \caption{ \label{fig:dog_holes}Examples of partial correspondence on the dog shape from the SHREC'16 Partial (holes) dataset. First row: correspondence produced by ACNN. Corresponding points are shown in similar color. Reference shape is shown on the left. 
   Second and third rows: pointwise geodesic error (in $\%$ of geodesic diameter) of the ACNN and RF correspondence, respectively. For visualization clarity, the error values are saturated at $10\%$ of the geodesic diameter. Hot colors correspond to large errors.}
\end{figure*}

\paragraph*{Holes.}
Figure~\ref{fig:shrec_holes} compares the performance of different partial matching methods on the SHREC'16 Partial (holes) dataset. In this setting as well, ACNN outperforms other approaches with a significant margin. 
Figure~\ref{fig:dog_holes} (top) shows examples of partial correspondence on the dog shape as well as the pointwise geodesic error (bottom). 

%
%
%

\section{Conclusions}
\label{sec:concl}

We presented Anisotropic CNN, a new framework generalizing convolutional neural networks to non-Euclidean domains, allowing to perform deep learning on geometric data. Our work follows the very recent trend in bringing machine learning methods to computer graphics and geometry processing applications, and is currently the most generic intrinsic CNN model. 
Our experiments show that ACNN outperforms previously proposed intrinsic CNN models, as well as additional state-of-the-art methods in the shape correspondence application in challenging settings. 
Being a generic model, ACNN can be used for many other applications. 
We believe it would be useful for the computer graphics and geometry processing community, and will release the code. 

\section*{Acknowledgment} 
This research was supported by the ERC Starting Grant No. 307047 (COMET), a Google Faculty Research Award, and Nvidia equipment grant.

\bibliographystyle{eg-alpha}
\bibliographystyle{eg-alpha-doi}
\bibliography{./biblio,./biblio2,./intro,./refs,./biblio3}

\end{document}